\documentclass[10pt,twocolumn,letterpaper]{article}

\usepackage{iccv}
\usepackage{times}
\usepackage{epsfig}
\usepackage{graphicx}
\usepackage{amsmath}
\usepackage{amssymb}

\usepackage{booktabs}
\usepackage{comment}
\usepackage{bm}
\usepackage{url}
\usepackage[accsupp]{axessibility} 
\newcommand{\cRisa}[1]{{\color{black}{#1}}} 
\newcommand{\cRio}[1]{{\color{black}{#1}}} %

\usepackage[pagebackref,breaklinks,colorlinks]{hyperref}
\iccvfinalcopy 

\def\iccvPaperID{****} 

\ificcvfinal\fi

\begin{document}

\title{SegRCDB: Semantic Segmentation via Formula-Driven Supervised Learning}

\author{Risa Shinoda\textsuperscript{\rm 1},
Ryo Hayamizu\textsuperscript{\rm 1},
Kodai Nakashima\textsuperscript{\rm 1},
Nakamasa Inoue\textsuperscript{\rm 1,2},
Rio Yokota\textsuperscript{\rm 1,2},
Hirokatsu Kataoka\textsuperscript{\rm 1}\\
\textsuperscript{\rm 1}National Institute of Advanced Industrial Science and Technology (AIST)\\
\textsuperscript{\rm 2}Tokyo Institute of Technology}

\maketitle
\ificcvfinal\thispagestyle{empty}\fi

\begin{abstract}
Pre-training is a strong strategy for enhancing visual models to efficiently train them with a limited number of labeled images. In semantic segmentation, creating annotation masks requires an intensive amount of labor and time, and therefore, a large-scale pre-training dataset with semantic labels is quite difficult to construct. Moreover, what matters in semantic segmentation pre-training has not been fully investigated.
In this paper, we propose the Segmentation Radial Contour DataBase (SegRCDB), which for the first time applies formula-driven supervised learning for semantic segmentation. SegRCDB enables pre-training for semantic segmentation without real images or any manual semantic labels.
SegRCDB is based on insights about what is important in pre-training for semantic segmentation and allows efficient pre-training.
Pre-training with SegRCDB achieved higher mIoU than the pre-training with COCO-Stuff for fine-tuning on ADE-20k and Cityscapes with the same number of training images.
SegRCDB has a high potential to contribute to semantic segmentation pre-training and investigation by enabling the creation of large datasets without manual annotation. The SegRCDB dataset will be released under a license that allows research and commercial use. Code is available at: \url{https://github.com/dahlian00/SegRCDB}
\end{abstract}

\section{Introduction}
\label{sec:intro}
Preparing a semantic segmentation dataset requires pixel-level, dense annotation, and therefore, creating a fully annotated dataset incurs a huge amount of effort. 
For a dataset such as Cityscapes~\cite{Cordts2016Cityscapes}, around 90 minutes per image is required for pixel-level annotation.
This makes it difficult to create a large semantic segmentation dataset.

To perform training with a limited dataset, using a pre-trained model with a large-scale image dataset is a promising method for enhancing network performance in terms of recognition accuracy. 
The use of a model pre-trained on a larger-scale image dataset has become a standard approach.
Undoubtedly, ImageNet~\cite{5206848} is one of the de-facto-standard datasets, even in the semantic segmentation field. However, ethical issues have been reported in terms of dataset biases and privacy violations~\cite{Prabhu2021LargeID,Yang_2020,yang2022faceobfuscation}. 
There have also been reports~\cite{zhao2021captionbias,wang2019iccv} that one of the most frequently used segmentation datasets (Microsoft COCO~\cite{https://doi.org/10.48550/arxiv.1405.0312}) also raises concerns regarding transfer learning due to ethical issues. 

\begin{figure}
\centering
\includegraphics[width=8cm]{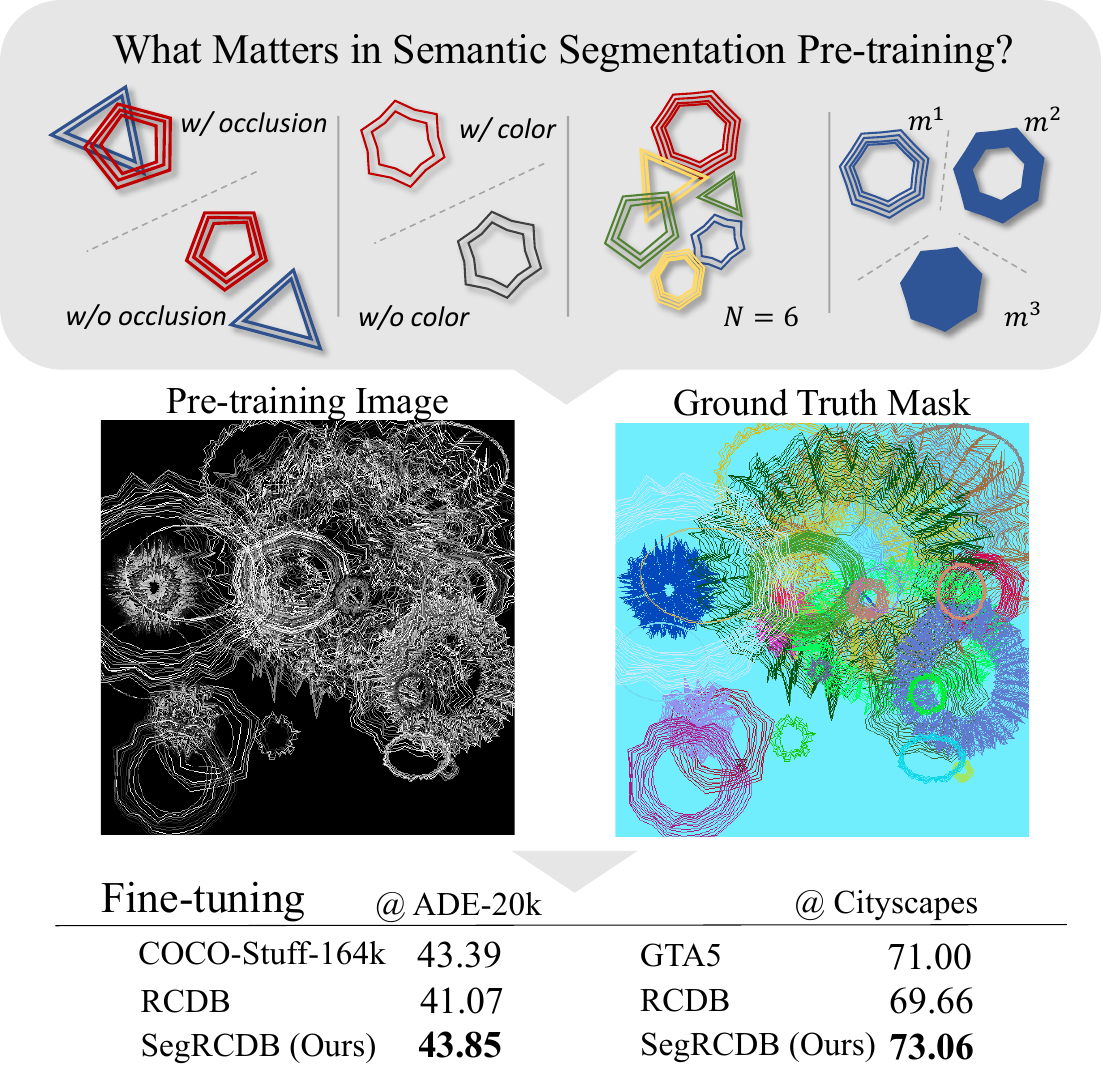}
\caption{\textbf{Segmentation Radial Contour DataBase (SegRCDB).} We developed the first formula-driven supervised learning (FDSL) method for semantic segmentation. We created a precise pixel-wise ground truth mask without manual effort.}
\label{figA}
\vspace{-0pt}
\end{figure}

To overcome these technical (manual annotation) and ethical (fairness and dataset transparency) problems, a synthetic dataset can be implemented to construct a pre-trained segmentation model. Synthetic datasets for semantic segmentation have been created for specialized use in certain domains~\cite{Richter_2016_ECCV, Ros_2016_CVPR,8237554}. However, even though the annotation time is reduced, McCormac~\textit{et al.} reported that it took up to approximately one month using 4-12 GPUs to produce a 5M-image dataset~\cite{8237554}. For automatic dataset creation, we must define a 3D scene design that includes object categories and positional content, and consider the camera trajectory.
There is still potential for more efficient development of synthetic segmentation data that is applicable to general domains. 

However, there has not been enough research to determine what factors are effective in pre-training semantic segmentation.
It is not clear which dataset parameters are useful for pre-training semantic segmentation, such as the number of classes and the presence of occlusions.
Identifying these critical parameters will allow us to build a synthetic dataset that is more efficient for pre-training semantic segmentation.

 Formula-driven supervised learning (FDSL)~\cite{Kataoka_2022_IJCV} has been proposed and improved in the context of visual representation learning without real images.
 FDSL enables the automatic construction of large-scale image datasets through simultaneous image and label generation based on a simple mathematical formula.
 It was reported that the image dataset, which consists of complicated contours (e.g., Radial Contour DataBase; RCDB), allows effective image classification~\cite{Kataoka_2022_CVPR}.
 The object contours are mainly captured in the pre-training phase, and the visual representation is transferred to determine object categories in a real image.
 The FDSL framework has not been applied to acquiring a visual representation for semantic segmentation. 

 In this paper, we propose the first formula-driven supervised learning for semantic segmentation and create
 the SegRCDB dataset.
 We assume that a model pre-trained using radial contours can further improve the recognition ability of semantic segmentation since semantic masks are assigned to radial contour areas.  
To take advantage of making image patterns and ground truth masks based on a simple equation without real-image collection or manual annotation, we thoroughly investigated the effectiveness of different configurations (e.g., occlusion between objects, types of mask annotation) in pre-training.
The SegRCDB will contribute to semantic segmentation tasks and reduce the time and effort required to create ground truths for large-scale image datasets.

The main contributions of this study are as follows:
\begin{itemize}
    \item We developed the first FDSL method for semantic segmentation. 
    We created a precise pixel-wise mask (Figure~\ref{figA}) without any manual effort. 
    \item We investigated what elements are effective for improving the accuracy in pre-training for semantic segmentation, and created SegRCDB based on the results.
    \item  Our SegRCDB has great potential for effectively pre-training a semantic segmentation model based on a massive amount of pixel-level ground truth. The proposed method performed better than the COCO-Stuff-164k baseline (e.g., 43.39 vs. 43.85 mIoU on ADE-20k) and the GTA5 baseline (e.g., 71.00 vs. 73.06 mIoU on Cityscapes).
\end{itemize}
\section{Related Work}
\label{sec:relatedwork} 
Here, we will limit the discussion to studies that are closely related to our proposed method.

\noindent \textbf{Semantic segmentation.} 
The earlier semantic segmentation models were implemented on CNN backbones and their heads ~\cite{7298965, https://doi.org/10.48550/arxiv.1505.07293, https://doi.org/10.48550/arxiv.1505.04597, 10.1007/978-3-030-01234-2_49, https://doi.org/10.48550/arxiv.1703.06870}. In semantic segmentation tasks, we must effectively acquire context information and perform precise labeling for a whole image. This property is inherited by Transformer models~\cite{vaswani2017attention, xie2021segformer,jain2021semask, liu2021Swin}. In particular, Swin Transformer~\cite{liu2021Swin} is frequently used as a strong baseline model due to its ability to capture spatial features. We also employed the Swin Transformer model in the experiments described in this paper. 

\noindent \textbf{Automatic/semi-automatic segmentation pre-training.}
One way to reduce annotation costs in semantic segmentation is by generating synthetic images and/or assigning ground truth masks. Previous studies have shown the effectiveness of increasing the ground truth by a pseudo-labeling technique~\cite{Zou_2019_ICCV, zou2018unsupervised,shin2020two,zheng2021rectifying,Li2019BidirectionalLF}, active learning~\cite{xie2022ripu,9156651,Qiao_Zhu_Long_Zhang_Wang_Du_Yang_2022} or by creating synthetic data and ground truth labels~\cite{semanticGAN}. However, these techniques may produce a biased dataset with the dominant class or images when real images are used.  Synthetic datasets also contribute to pre-training methods such as GTA5~\cite{Richter_2016_ECCV} and SYNTHIA~\cite{Ros_2016_CVPR}. Large-scale synthetic datasets for semantic segmentation decrease the manual annotation time, but they still need to create and define models and environments.

\noindent \textbf{Ethical issues with image datasets.} Large-scale datasets have been reported to have ethical issues. For semantic segmentation, COCO-Stuff~\cite{https://doi.org/10.48550/arxiv.1612.03716} is one of the largest datasets, and it contains 164k images from the COCO dataset. However, it has been reported that there are biases in the COCO dataset in relation to gender and race~\cite{zhao2021captionbias,wang2019iccv}. Even the de-facto standard ImageNet dataset is associated with fairness and privacy problems due to biased inputs and human-related images~\cite{Prabhu2021LargeID,Yang_2020}. 

\noindent \textbf{FDSL.} 
To overcome dataset-related problems with real images, formula-driven supervised learning (FDSL) pre-training methods have been developed~\cite{Kataoka_2022_IJCV}.
In FDSL, a simple mathematical formula (e.g., fractals~\cite{Kataoka_2022_IJCV,anderson2022improving,nakashima2022can,Kataoka_2022_CVPR}, tiling patterns~\cite{kataoka2021formula}, and contours~\cite{Kataoka_2022_CVPR}) allows automatic generation of image patterns and their training labels at the same time.
This framework makes it possible to create large-scale pre-training datasets without any real images. Recent studies have verified that a mathematical formula can be used to pre-train models for video~\cite{kataoka2022stinitialization} and 3D objects~\cite{yamada2021iros,yamada2022point}.
Moreover, Kataoka~\textit{et al.}~\cite{Kataoka_2022_CVPR} revealed that an image dataset consisting of complex contours (e.g., Radial Contour DataBase; RCDB) was very effective at pre-training models for image classification.
We believe that RCDB is also effective for semantic segmentation because a model pre-trained using RCDB can acquire accurate object contours.
Past studies (e.g.,~\cite{Bertasius2016ssboundary,takikawa2019gatedscnn}) have reported that recognizing boundaries is important for semantic segmentation, and is also significant in segmentation pre-training.

\def\figB{
\begin{figure}
\centering
\includegraphics[width=8.1cm]{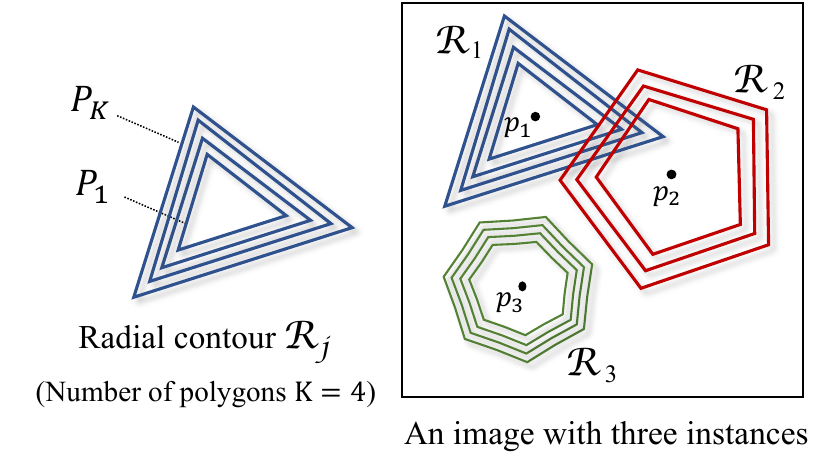}
\caption{{\bf Radial contours.} An image $\bm{x}_{i}$ is composed of multiple radial contours. The radial contours $\mathcal{R}_{j}$ are placed at positions $\bm{p}_{j}$. Each radial contour is an object made by superimposing polygons $\{P_{k}\}_{k=1}^{K}$.}
\label{figB}
\vspace{-15pt}
\end{figure}
}

\def\figC{
\begin{figure*}
\centering
\includegraphics[width=17cm]{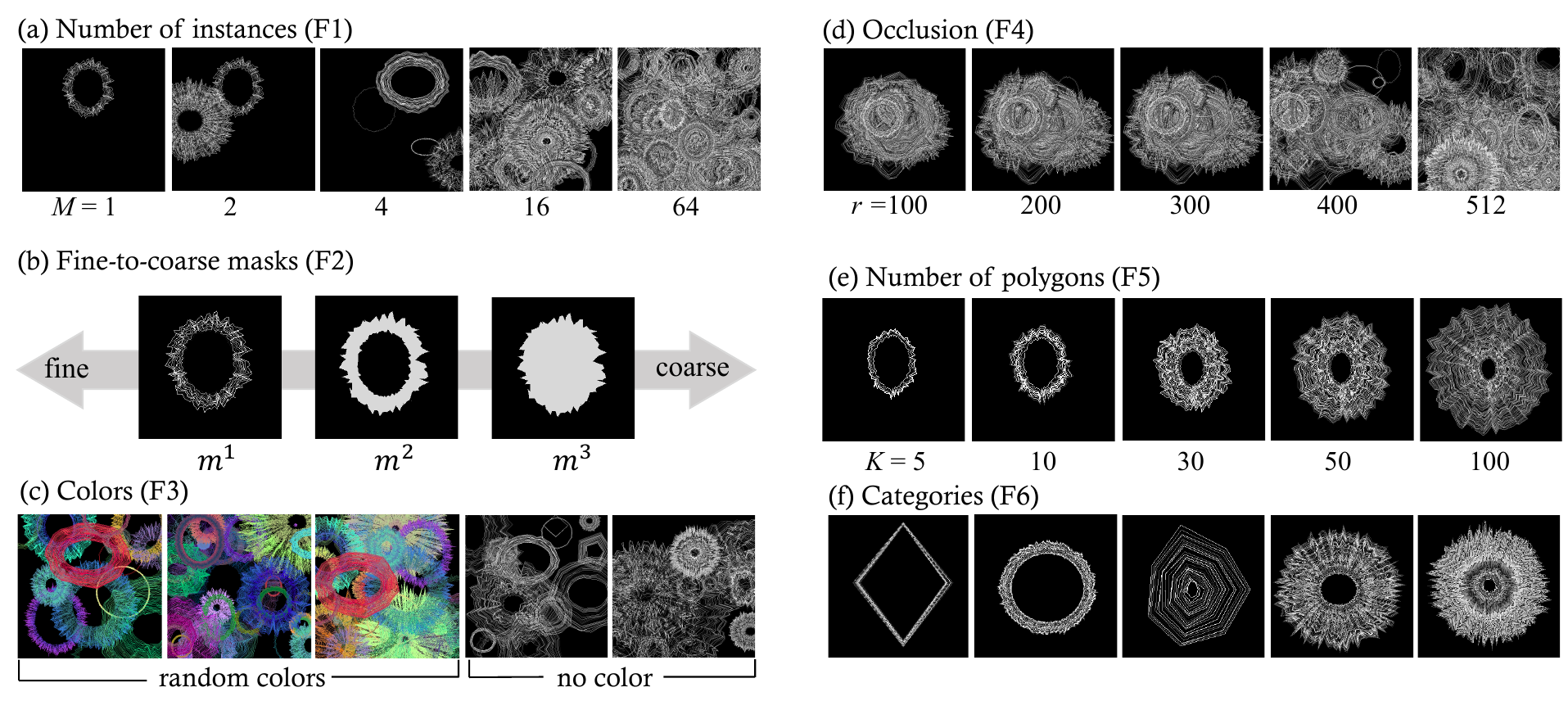}
\vspace{-10pt}
\caption{{\bf Examples of images used in the investigation.}}
\label{figC}
\vspace{-15pt}
\end{figure*}
}

\section{Investigation Policy for Semantic Segmentation Pre-training}
\label{sec:investigation_policy}
In this section, we present a framework for investigating what is most important for improving the pre-training accuracy for semantic segmentation without relying on a collection of real images.
We first describe our investigation policy involving seven factors to be parameterized when synthesizing a dataset based on a formula-driven approach
and then propose a simple yet effective method for dataset synthesis as well as a pre-training method
that enables us to empirically investigate the effectiveness of each factor. For synthetic image generation using a formula-driven approach, we can create clear patterns and perfect semantic labels to easily analyze disassembled components in pre-training for semantic segmentation. We are the first to propose a pre-training method for semantic segmentation with the formula-driven supervised method.

\noindent {\bf Setting.}
This paper considers the setting where a training dataset for semantic segmentation consists of pairs of images and ground truth masks.
Specifically, a dataset is given by $\mathcal{D} = \{(\bm{x}_{i}, \bm{m}_{i})\}_{i=1}^{N}$ where $\bm{x}_{i}$ is an image, $\bm{m}_{i}$ is a ground truth mask, and $N$ is the number of images.
Each pixel in a mask $\bm{m}_{i}$ represents a category label $c \in \{0, 1, 2, \cdots, C\}$ for its corresponding pixel at $\bm{x}_{i}$, where $C$ is the number of object categories. $c=0$ is used to indicate the background.

We assume that an image $\bm{x}_{i}$ is composed of $M_{i}$ individual object instances $\{\bm{o}_{j}\}_{j=1}^{M_{i}}$ and a background, where the instances are ordered from the back to the front, {\it i.e.}, $\bm{o}_{1}$ is the backmost instance and $\bm{o}_{M_{i}}$ is the frontmost instance.
Each instance has a category label $c_{j}$ and a binary mask $\bm{b}_{j} \in \{0, 1\}^{W \times H}$ that describes the pixel-level appearance of the instance, where $W$ and $H$ are the image width and height, respectively.
Under this assumption, the mask $\bm{m}_{i}$ is obtained by
\begin{align}
\label{eq:mask}
[\bm{m}_{i}]_{p,q}
=
c_{j^{*}},~~
j^{*} = \max(\{j : [\bm{b}_{j}]_{p,q} = 1\} \cup \{0\})
\end{align}
where $[\cdot]_{p,q}$ indicates the element at pixel position $(p, q)$ within the image.
The $\max$ operation in Equation~(\ref{eq:mask}) is used to consider partial occlusion of instances, so that only the frontmost category label is observable.
Note that $c_{0} = 0$ is used for the background.

\figB
\figC

\noindent {\bf Investigation policy.}
In the above setting, the number of training images $N$ is often important for improving the pre-training performance, and many previous studies have shown the effectiveness of large-scale datasets of real images~\cite{https://doi.org/10.48550/arxiv.1405.0312}.
However, the effects of other factors such as the number of instances $M_{i}$ have been rarely investigated.
When pre-training neural networks with synthesized images, what factors improve the performance?
To answer this question, we consider the following factors:
{
\setlength{\leftmargini}{22pt}
\begin{enumerate}
\setlength{\itemsep}{0pt}
\setlength{\parskip}{0pt}
\setlength{\itemindent}{0pt}
\setlength{\labelsep}{3pt}

\item[\bf (F1)]
\noindent {\bf Number of instances.}
How many instances should we have in one image? As the number of instances increases, the segmentation task becomes difficult to solve. We investigate the optimal task difficulty for pre-training in terms of the number of instances $M_{i}$.

\item[\bf (F2)]
\noindent {\bf Mask accuracy.}
How accurate should masks be? We investigate the effects of fine-to-coarse masks for $\bm{m}_{i}$.

\item[\bf (F3)]
\noindent {\bf Colors.}
Are colors necessary? We investigate whether the color channels of $\bm{x}_{i}$ help improve the results.

\item[\bf (F4)] \noindent {\bf Occlusion.}
Does partial occlusion of some instances boost performance?
We investigate how overlap between instances affects pre-training.

\item[{\bf (F5)}]{\bf Instance shapes.}
How complex should the instance shapes be?
We investigate the necessary complexity of the boundary shapes for instances $\bm{o}_{j}$
as well as the optimal line width and number of polygons in each instance for pre-training.

\item[\bf (F6)]
\noindent {\bf Number of categories.}
Do we need various categories? We investigate the importance of the number of categories $C$, which corresponds to the number of channels for the masks.

\item[\bf (F7)]
\noindent {\bf Number of images.}
Is increasing the number of images the only way to improve pre-training accuracy?
Finally, we compare the importance of increasing the number of images $N$ with the other factors.
\end{enumerate}
}

\def\datasetname{SegRCDB}
\section{\datasetname}
\label{sec:SegRCDB}

Based on the Section~\ref{sec:investigation_policy},
we propose {\datasetname},
a dataset of synthesized images and masks for pre-training semantic segmentation networks.
The dataset is designed to enable us to control the seven factors (F1)-(F7).
In the following, we describe how {\datasetname}, $\mathcal{D}_{\text{\datasetname}} = \{(\bm{x}_{i}, \bm{m}_{i})\}_{i=1}^{N}$ is constructed.

\noindent {\bf Instances.}
To control the complexity of instance shapes, we use {\it radial contours}, the polygonal objects proposed in \cite{Kataoka_2022_CVPR}, for object instances $\{\bm{o}_{j}\}_{j=1}^{M_{i}}$.
More specifically, an instance $\bm{o}_{j}$ is given by
\begin{align}
\bm{o}_{j} = (\mathcal{R}_{j}, \bm{p}_{j}, c_{j})
\end{align}
where $\mathcal{R}_{j}$ is a radial contour, $\bm{p}_{j}$ is the position in the image, and $c_{j}$ is the category label.
Here, a radial contour $\mathcal{R}_{j}
\subset \mathbb{R}^{2}$ is a 2D object made by superimposing polygons as $\mathcal{R}_{j} = \cup_{k=1}^{K} P_{k}$ where $P_{k}$ is a skeleton polygon ({\it e.g.}, a triangle) and $K$ is the number of polygons.
As shown in Figure~\ref{figB}, an image $\bm{x}_{i}$ is composed of $M_{i}$ radical contours.
Please refer to \cite{Kataoka_2022_CVPR} for the detailed definition of each instance $R_{k}$.

\noindent {\bf Number of instances.}
The number of instances per image $M$ (F1) is treated as a hyper-parameter by assuming $M_{i} = M$ for all $i = 1, 2, \cdots, N$.
In the experiments, $M$ is chosen from $\{1, 2, 4, 8, 16, 32, 64\}$.
Some example images are shown in Figure~\ref{figC}a.

\noindent {\bf Masks.}
We introduce three types of masks in a fine-to-coarse manner for the investigation of (F2).
The first mask $\bm{m}^{1}_{j}$ is the finest mask, which is the mask over skeleton lines obtained by the following two steps.
First, binary masks $\bm{b}_{j}$ are synthesized by rendering a radial contour $\mathcal{R}_{j}$ as a binary image, {\it i.e.}, by rendering white lines on a black background using the same method as that for synthesizing color images.
Second, masks are computed using Equation~(\ref{eq:mask}).
The second mask $\bm{m}^{2}_{j}$  is a ring mask obtained by filling the region between the first polygon $R_{1}$ and the final polygon $R_{K}$ with a value of $1$.
The third mask $\bm{m}^{3}_{j}$ is the coarsest mask obtained by filling the region inside $R_{K}$ with the value of $1$.
Examples of these masks are shown in Figure~\ref{figC}b.

\noindent {\bf Colors.}
To render instances $\{\bm{o}_{j}\}_{j}^{M_{i}}$ into an image $\bm{x}_{i}$,
we introduce coloring methods.
Specifically, we use either of two methods for the investigation of (F3).
The first method attaches a random color to each instance $\bm{o}_{j}$, where RGB color values are sampled from the uniform distribution over $\{0, 1, \cdots, 255\}$.
The second method ignores colors and renders all instances in white. The background color is fixed to black for both methods. Example images are shown in Figure~\ref{figC}c.

\noindent {\bf Occlusion.}
Occlusion is introduced for the investigation of (F4). We generate occlusion by changing the position of polygons. The center position of the polygon is adjusted to \{100, 200, 300, 400, 512\} pixels, centered on the image. Some images with different occlusion levels are shown in Figure~\ref{figC}d. The smaller the value of $r$, the more polygons are concentrated in the center and the more occlusions occur.

\noindent {\bf Instance shapes.} The number of polygons per instance $K$ (F5) is treated as a hyper-parameter. Some images with different numbers of polygons are shown in Figure~\ref{figC}e. The line width $d$ is chosen from \{1, 2, 3\} pixels.

\noindent {\bf Categories.}
The category (F6) is determined by the number of vertices, the radius, a resizing factor, and Perlin noise. Please refer
to~\cite{Kataoka_2022_CVPR} for details. The parameter of the number of categories $C$ is chosen from  \{64, 128, 255, 500\}. Some example images are shown in Figure~\ref{figC}f.

\noindent {\bf Number of images.}
The number of images $N$ (F7) is also treated as a hyper-parameter. 

\def\figD{
\begin{figure}[t]
\centering
\includegraphics[width=7cm]{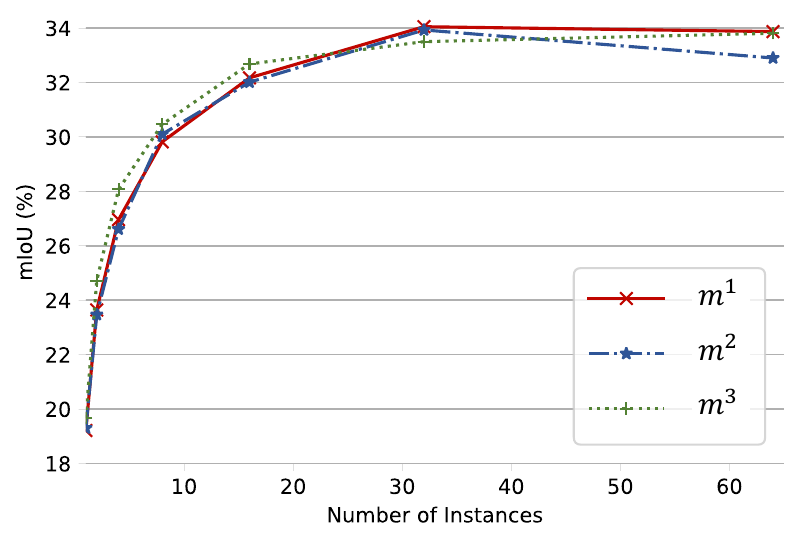}
\caption{{\bf Number of instances and mask types.} These experiments are related to factors (F1) and (F2).}
\label{figD}
\vspace{-15pt}
\end{figure}
}

\def\figE{
\begin{figure}
\centering
\includegraphics[width=8cm]{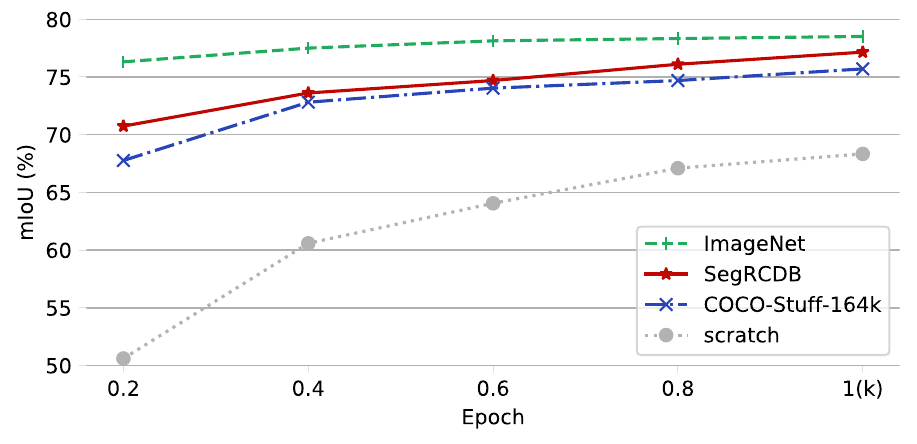}
\caption{{\bf mIoU transition during fine-tuning on Cityscapes.}}
\label{figE}
\vspace{-15pt}
\end{figure}
}
\section{Experiments}

In this section, we evaluate the pre-training effects on SegRCDB from multiple aspects. Especially, we follow the investigation policy described in Section~\ref{sec:investigation_policy}. At the beginning of this section, we present results for factors (F1) to (F7). 
Based on the investigation, we build SegRCDB with the best parameters for each element and  evaluate its performance compared to representative datasets.

\begin{table}
  \centering
  \caption{Baseline parameter set (see Section~\ref{sec:SegRCDB} for detailed parameter descriptions).}
  \begin{tabular}{lc}
    \toprule
     Baseline parameter&\\
    \midrule
        Line width ($d$) & $1.0$ \\
        Number of polygons ($K$) & \{$1$, $2$, $3$,...,$50$\}\\
        Occlusion ($r$)  & $512$\\
        Color & Grayscale\\
        Number of categories ($C$) & $255$ \\
        Number of images ($N$) & $20$k\\
    \bottomrule
  \end{tabular}
  \label{benchmark}
\end{table}

\subsection{Implementation Details}
\label{sec:investigate_setting}
\noindent{\bf Model.} In the segmentation pre-training experiments, we used Swin Transformer base model~\cite{liu2021Swin} as a backbone, and UPerNet~\cite{xiao2018unified} as the entire architecture. For implementation, we use codes and models in MMSegmentation~\cite{mmseg2020}. 

\noindent {\bf Loss function.} 
We follow the official implementation~\cite{liu2021Swin}.
The backbone part was pre-trained with the AdamW optimizer~\cite{Loshchilov2019DecoupledWD} with a weight decay of 0.01. For the training of UPerNet, we adapted cross-entropy loss. 

\noindent {\bf Learning schedule.} 
The pre-training length is adjusted to 300 epochs by following the FDSL paper~\cite{Kataoka_2022_CVPR}. Additionally, the fine-tuning length was 60 epochs. Through the preliminary study, the batch size was set as 32.

\noindent {\bf Dataset for pre-training and fine-tuning.} 
For comparison, we used ADE-20k~\cite{zhou2017scene} as a standard semantic segmentation dataset. This dataset contains 150 categories from daily living environments. The dataset is divided into 20,210 images for training, and 2,000 images for validation. In examining the effect of parameters, baseline parameters were set as shown in Table~\ref{benchmark}. 
\figD
\subsection{Investigation Results}
\label{sec:investigation_result}

\begin{table*}[t]
\begin{tabular}{ccc}
\hspace{-10pt}
\begin{minipage}[t]{0.3\hsize}
  \caption{(F3) Grayscale vs. color.} 
\begin{center}
\scalebox{1.0}{
  \begin{tabular}{l|cc}
    \toprule
      Type & Gray &  Color\\
     \midrule
       mIoU  & \textbf{34.05} & 28.49 \\
    \bottomrule
  \end{tabular}
  \label{tab:color}
  }
\end{center}
\end{minipage}
\hspace{-20pt}
\begin{minipage}[t]{0.45\hsize}
  \caption{(F4) Occlusion.} 
\begin{center}
\scalebox{1.0}{
  \begin{tabular}{c|ccccc}
    \toprule
     $r$ &  100 & 200 & 300 & 400 & 512\\
     \midrule
     mIoU & 30.62 & 31.87  & 33.20 & \textbf{34.12} & 34.05\\
    \bottomrule
  \end{tabular}
  \label{tab:occulusion}
 }
\end{center}
  \end{minipage}
  \begin{minipage}[t]{0.29\hsize}
\caption{(F5) Number of polygons.} 
 \vspace{-20pt}
\begin{center}
\scalebox{1.0}{
  \begin{tabular}{c|ccc}\toprule[0.8pt]
      $K$ & 1-25 & 26-50 & 1-50\\
     \midrule
       mIoU & \textbf{34.12} & 33.72  & 34.05 \\
    \bottomrule
  \end{tabular}
  \label{tab:line}
 }
\end{center}
  \end{minipage}
     \end{tabular}
\end{table*}

\begin{table*}[t]
\begin{tabular}{ccc}
\hspace{-10pt}
  \begin{minipage}[t]{0.31\hsize}
  \caption{(F5) Line width.} 
\begin{center}
\scalebox{1.0}{
  \begin{tabular}{c|ccc}\toprule[0.8pt]
      $d$ & 1px & 2px& 3px\\
     \midrule
      mIoU  &\textbf{34.05} & 34.03& 34.03 \\
    \bottomrule
  \end{tabular}
  \label{tab:width}
  }
\end{center}
  \end{minipage}
\hspace{-5pt}
 \begin{minipage}[t]{0.38\hsize}
\caption{(F6) Number of categories.} 
 \begin{center}
\scalebox{1.0}{
  \begin{tabular}{c|cccc}\toprule[0.8pt]
      $C$ & 64 & 128  & 255 & 500\\
     \midrule 
      mIoU  & 31.67& 33.62 & \textbf{34.05} & 31.58\\
    \bottomrule[0.8pt]
  \end{tabular}
  \label{tab:numof_category}
}
\end{center}
  \end{minipage}
 \begin{minipage}[t]{0.31\hsize}
\caption{(F7) Number of images.} 
 \begin{center}
\scalebox{1.0}{
  \begin{tabular}{c|ccc}\toprule[0.8pt]
      $N$ & 20k & 40k & 80k \\
     \midrule 
      mIoU  & 41.31& 41.86& \textbf{42.25}\\
    \bottomrule
  \end{tabular}
  \label{tab:numof_image}
   }
\end{center}
  \end{minipage}
     \end{tabular}
\end{table*}

\noindent {\bf (F1/F2) Number of instances, and mask accuracy.}
The experimental results for the number of instances per image and the mask type are both shown in Figure~\ref{figD}. We assigned \{2, 4, 8, 16, 32, 64\} for the number of instances and \{$\bm{m}^{1}$, $\bm{m}^{2}$, $\bm{m}^{3}$\} for the mask type.
As shown in Figure~\ref{figD}, the more instances for an image, the better the accuracy that can be achieved, up to 32 instances.
With regard to the mask type, the most pixel-wise annotation $\bm{m}^{1}$ achieves the highest results with 32 instances. Here, pixel-wise annotation is extremely difficult for human annotators in recently reported work~\cite{qin2022}. Although humans always find annotation very difficult, our SegRCDB can easily create precise segmentation masks with semantic labels, due to the formula-driven approach using a simple equation.  
Hereafter, 32 instances per image and a ${m}^{1}$ mask type are used.

\noindent {\bf (F3) Colors.} Table~\ref{tab:color} shows the results for color type. We compared the grayscale dataset with a dataset randomly colored by class number. The grayscale dataset achieved a better result, indicating that focusing on object contours produces better results than learning colors for each class. As in a previous paper~\cite{Kataoka_2022_CVPR}, object contours were found to be more important than color in transformer-based self-attention approaches, including Swin Transformer backbone and UPerNet head.

\noindent {\bf (F4) Occlusion.} Table~\ref{tab:occulusion} shows how occlusion affects the experimental results. The parameter $r$ represents the range where the polygon is drawn. After careful investigation, we assigned the parameters as \{200, 300, 400, 512\} for $r$. The highest mIoU was achieved for 400 for the fine-tuned dataset. This indicates that an appropriate amount of occlusions can contribute to better results. We can adjust the task difficulty by using the $r$ parameter related to the amount of occlusion inside of the image. 

\noindent {\bf (F5) Instance shapes}. We investigated the effect of the complexity of the instance shape, by varying the number of polygons and the line width. We set the number of polygons as \{1--25, 26--50, 1--50\}, and the line width as \{1, 2, 3\} pixels. The effects of the number of polygons and the line width are shown in Table~\ref{tab:line} and \ref{tab:width}, respectively. Fewer polygons and thinner lines produced slightly better results than those of other configurations. Thinner lines lead to fine annotations. 

\noindent {\bf (F6) Number of categories}. The experimental results for different numbers of categories in the radial contours are shown in Table~\ref{tab:numof_category}. We employed \{64, 128, 255, 500\} in pre-training.
Almost all models adopt an 8-bit mask input, so we tried the maximum number 255 (category 256 is used for data augmentation). Using the 16-bit mask image, we also created 500 categories.
Based on the results, the 255 class shows the highest results. Hereafter, the number of categories is set at 255.

\noindent {\bf (F7) Number of images.} The pre-training effects of the number of images are shown in Table~\ref{tab:numof_image}. Fine-tuning is set to 120 epochs to allow for convergence. As the number of images increases, the mIoU value improves. This confirms the intuition that large-scale datasets can be effective in pre-training for semantic segmentation.
In Section~\ref{sec:semseg_comparison},  we adopt 118k images for SegRCDB, which \cRio{has the same number of training images} as the COCO-Stuff-164k for the comparison of pre-training with semantic segmentation datasets.

\subsection{Comparison}

\begin{table*}[t]
\begin{minipage}[t]{.5\hsize}
  \centering
  \caption{Comparison of pre-training with semantic segmentation datasets. The best and second-best values for each fine-tuning dataset are in underlined bold and bold, respectively.}
  \begin{tabular}{lccccc}
    \toprule
        &  &\multicolumn{2}{c}{ADE-20k} & \multicolumn{2}{c}{Cityscapes} \\
        Pre-training & \#Img &  mIoU & mAcc & mIoU & mAcc \\
    \midrule
        Scratch&-&31.40 &41.02 &54.65&62.89 \\
        ADE-20k& 20k & - &-&68.46 &77.13\\
        GTA5& 25k &39.31&49.79& 71.00&79.31 \\
        COCO-Stuff& 118k &\textbf{43.39} &\textbf{54.41} &\textbf{72.21}&\textbf{80.62} \\
        SegRCDB&118k & \underline{\textbf{43.85}}&\underline{\textbf{54.98}} & \underline{\textbf{73.06}} &\underline{\textbf{81.59}} \\
    \bottomrule
    \end{tabular}
  \label{segment_comparison}
  \vspace{-5pt}
\end{minipage}
\begin{minipage}[t]{.48\hsize}
 \centering
 \caption{Comparison with backbone pre-training. The best and second-best values for each fine-tuning dataset are in underlined bold and bold, respectively.} 
    \begin{tabular}{lccccc}
      \toprule
        &  &\multicolumn{2}{c}{ADE-20k} & \multicolumn{2}{c}{Cityscapes} \\
        Pre-training & \#Img &  mIoU & mAcc & mIoU & mAcc \\
        \bottomrule[0.5pt]
        Scratch & -& 31.40 & 41.02 & 54.65 & 62.89 \\
        ImageNet& 1.28M & \underline{\textbf{46.37}} & \underline{\textbf{57.11}} & \underline{\textbf{75.26}}& \underline{\textbf{83.60}} \\
        ExFractalDB&1M&  40.96 & 52.13  & 68.93 & 77.96\\
        RCDB&1M &  41.07 & 51.89 &  69.66 & 78.35  \\
        SegRCDB&118k & \textbf{43.85} & \textbf{54.98} & \textbf{73.06} & \textbf{81.59}    \\
        \bottomrule
 \end{tabular}
\label{backbone_compare}
\end{minipage}
\end{table*}

In this section, we investigate the pre-training effects on SegRCDB, the first semantic segmentation dataset created in the framework of FDSL. To evaluate the effectiveness of various factors, we compared the results of pre-training with semantic segmentation datasets and backbone pre-training with large datasets. 

\noindent \textbf{Supervised pre-training for semantic segmentation datasets.}
\label{sec:semseg_comparison}
We compared the pre-training effects of our SegRCDB with representative semantic segmentation datasets based on supervised learning.

\begin{itemize}
\item \textbf{COCO-Stuff~\cite{https://doi.org/10.48550/arxiv.1612.03716}} contains labeled images from the COCO dataset with pixel-level object annotations. The dataset has 171 categories and 118k training images.
\item \textbf{Cityscapes~\cite{Cordts2016Cityscapes}} contains labeled images with 19 categories captured from street scenes. The dataset is divided into 2,975 images for the training set, and 500 images for the validation set. 
\item \textbf{GTA5~\cite{Richter_2016_ECCV}} contains synthetic street images rendered from the GTA5 video game. The dataset was annotated into 19 categories and contains only a training dataset for pre-training usage.
\item \textbf{ADE-20k~\cite{zhou2017scene}} is described in Section~\ref{sec:investigate_setting}. 
\end{itemize}

All datasets were pre-trained by Swin Transformer base model~\cite{liu2021Swin} for the backbone and UPerNet~\cite{xiao2018unified} for the entire architecture.
The entire UPerNet, including the backbone and head, is pre-trained on these semantic segmentation datasets and the pre-training effects are compared.
We used the
AdamW optimizer~\cite{Loshchilov2019DecoupledWD} with a weight decay of 0.1 following the official implementation. All datasets were cropped to produce input images of $512 \times 512$ pixels.  The pre-training length was 300 epochs, and the fine-tuning length was 150 epochs. 
The batch size was set to 64 for pre-training, and 16 for fine-tuning.  All datasets were fine-tuned with a backbone learning rate of 0.0005, and a weight decay of 0.1. 

Table~\ref{segment_comparison} shows comparisons with supervised pre-training methods for fine-tuning for ADE-20k and Cityscapes validation datasets.
According to the results, our SegRCDB achieves the highest score.
SegRCDB shows superior results to COCO-Stuff dataset even with the same amount of training data. Since SegRCDB selected effective parameters for pre-training, it outperformed the current standard pre-training dataset for semantic segmentation.

\noindent \textbf{Backbone pre-training for semantic segmentation}.
Here, we compare our SegRCDB with backbone pre-training before semantic segmentation fine-tuning. RCDB-1k and ExFractalDB-1k are the FDSL classification datasets, containing 1.0 million images.

For ImageNet-1k, RCDB-1k, and ExFractalDB-1k training, we followed the official settings of Swin Transformer~\cite{liu2021Swin}, except for the augmentation process. Augmentation is adjusted to the settings in MMSegmentation for all datasets. The pre-training length is 300 epochs, and the fine-tuning length is the 150 epochs. For the SegRCDB, the experiment settings are the same as in Table~\ref{segment_comparison}. SegRCDB is pre-trained on the UPerNet backbone and head, while ImageNet-1k, RCDB-1k, and ExFractalDB-1k datasets are trained on the backbone only.

Table~\ref{backbone_compare} shows experimental results comparing the effects of backbone pre-training with large-scale datasets and SegRCDB's pre-training. For ADE-20k and Cityscapes fine-tuning, ImageNet shows the highest mIoU. Among FDSL methods, SegRCDB achieves the highest scores. SegRCDB contains only 118k images; however, it can surpass other FDSL backbone pre-training methods with 1 million images. 
This indicates that the FDSL method, which has been considered effective for classification problems so far, has been successfully applied to semantic segmentation in SegRCDB.

\subsection{Explorative Study}
\label{sec:expolarative}
Additional experiments were conducted to further investigate the performance of SegRCDB. 

\noindent \textbf{mIoU transition during fine-tuning.}
\figE
Models trained by formula-driven supervised learning might require a longer fine-tuning epoch to acquire a real-image representation. Since previous studies of FDSL have employed 1k epoch fine-tuning~\cite{Kataoka_2022_CVPR, nakashima2022can},  we also conduct 1k epoch fine-tuning using SegRCDB of 118k images.
The results of 1k epoch fine-tuning on Cityscapes are shown in Figure~\ref{figE}. This shows that the pre-trained model with SegRCDB shows higher mIoU than from scratch and COCO-Stuff from the early stages of fine-tuning. The gap in mIoU between SegRCDB and ImageNet is wider in the early stages of fine-tuning but becomes narrower as the learning progresses.

\noindent \textbf{Large-scale SegRCDB.}
Compared to one instance per image in the conventional FDSL method, SegRCDB has 32 instances per image, which increases its pre-training effect.
We prepared a 1M scale SegRCDB to verify whether learning with a larger number of images would further enhance the pre-training effect.
Table~\ref{tab:image_size} shows the fine-tuning results for two different numbers of images of SegRCDB. SegRCDB with 1M images is trained for 150 epochs.  Other parameter settings are the same as Table~\ref{segment_comparison} and Table~\ref{backbone_compare}.
This indicates that fine-tuning on ADE-20k is more effective when the number of images is increased, while fine-tuning on Cityscapes is not.
It was also reported in previous studies that increasing the number of images does not improve the results for some fine-tuning datasets using the FDSL method~\cite{Kataoka_2022_CVPR, nakashima2022can}.

\noindent \textbf{CNN backbone architecture.}
We also investigated the performance of SegRCDB with different backbone networks. 
We used the ResNet-101~\cite{HeCVPR2016} convolutional backbone, as implemented in MMSegmentation~\cite{mmseg2020}. We used the same backbone model for both the pre-training and fine-tuning. Pre-training learning schedule was set to 300 epochs, while the fine-tuning was set to 150 epochs. We use 20k images of SegRCDB for training. Table~\ref{tab:backbone_compare} compares the backbone pre-training results for fine-tuning on ADE-20k and Cityscapes, where Swin-B shows higher results than those of the ResNet-101 model. As shown in previous work~\cite{Kataoka_2022_CVPR}, a radial contour shape is more effective for transformer architectures than for CNN architectures.

\noindent \textbf{Impact of annotation accuracy.}
In the semantic segmentation domain, it is often challenging to guarantee consistency between manually annotated ground truth masks. In this work, the FDSL approach ensured precisely labeled semantic annotations, thereby removing ambiguities and mistakes from annotations. In the dataset analysis of ADE-20k~\cite{zhou2017scene}, it is reported that on average only 82.4\% of pixels have the same annotation when re-annotated by the same annotator. Therefore, we deliberately shifted and inflated the pixel annotations on SegRCDB for pre-training. We varied the shift and inflation parameters in the range \{10, 30, 100, 300\} to emulate the statistics from the ADE-20k paper~\cite{zhou2017scene}. 
We conducted two experiments related to annotations: (a) shift, where we added varying degrees of noise to the object vertexes, and (b) inflation, which was obtained by enlarging the area enclosed by polygons.

Table~\ref{tab:annotation_analysis} shows the relationship between annotation precision and semantic segmentation performance in terms of mIoU. We used 20k training images for each trial. The learning schedule was set to 300 epochs for the pre-training stage, and to 150 epochs for fine-tuning on Cityscapes.
Our experiments show that shift annotation has a bigger impact on performance compared to inflation, lowering the mIoU score by  $6.54$ points when switching from 0 to 100 pixels' shift ($70.23 - 63.69$). In contrast, the mislabeled annotation caused by inflation appears less critical to the mIoU performance, with a decrease of 2.93 points up to 100 pixels. For a 300 pixel shift and inflation, the mIoU drops more significantly.

\subsection{Discussion and Limitations}
\begin{table}[t]
 \centering
 \caption{The comparison for the number of images. } 
    \begin{tabular}{lcccc}
      \toprule
          &\multicolumn{2}{c}{ADE-20k} & \multicolumn{2}{c}{Cityscapes} \\
         \#Img &  mIoU & mAcc & mIoU & mAcc \\
        \bottomrule[0.5pt]
         118k&43.85 &54.98 & 73.06 & 81.59    \\
       1M& 44.46 &55.67& 72.10&81.08  \\
        \bottomrule[0.8pt]
 \end{tabular}
\label{tab:image_size}
\end{table}

\begin{table}[t]
  \centering
  \caption{Comparison of backbone architecture.}
  \begin{tabular}{lcccc}
    \toprule
     & \multicolumn{2}{c}{ADE-20k}& \multicolumn{2}{c}{Cityscapes}\\
     Backbone   &mIoU&mAcc&mIoU&mAcc\\
    \midrule
       ResNet-101&39.56&51.48&66.74&75.31 \\
       Swin-B &41.51&52.58&70.23& 78.78 \\
    \bottomrule
  \end{tabular}
  \label{tab:backbone_compare}
  \vspace{-5pt}
\end{table}

We here summarize the findings obtained in the present study. According to the investigation described in Section~\ref{sec:investigation_result}, we confirmed that five factors are more important than other factors for semantic segmentation pre-training: (F1) number of instances, (F3) grayscale, (F4) occlusion, (F6) number of categories, and (F7) number of images.

\noindent \textbf{Number of instances (see also Figure~\ref{figD})}. The more instances for an image, the better accuracy is until saturation occurs at 32 instances. Increasing the number of instances in one image made the task more difficult, which was effective for pre-training.

\noindent \textbf{Grayscale representation (see also Table~\ref{tab:color}).} In relation to (F3), we confirmed that a grayscale representation is much better than a color image in object areas. The performance gap between grayscale and color was at 5.56 for ADE-20k. This suggests that the object categories should not be distinguished by color alone in a pre-training task. The color of objects may be an easy pre-training task, and cause the pre-trained model to be weak. Conversely, grayscale objects must be classified by means of object shapes using radial contours. This is inherited from the previous work~\cite{Kataoka_2022_CVPR}, and is also good for semantic segmentation.

\noindent \textbf{Occlusion (see also Table~\ref{tab:occulusion}).} From the results, occlusions among objects is important. Heavy occlusions can be solved in a pre-training task; however, for use as a hyper-parameter, there is an optimal occlusion amount. We confirmed that a value of $400$ is the most effective, and saturation occurs at $400$ because the value of $500$ produces almost the same level of mIoU on the ADE-20k dataset. The performance rate was increased by 3.5 points from minimal occlusions to more occlusion areas for these experimental settings.

\noindent \textbf{Number of categories and images (see also Tables~\ref{tab:numof_category} and \ref{tab:numof_image}).} In the investigations regarding factors (F6) and (F7), the numbers of categories and images are related to the pre-training effect. \cRisa{We confirmed that 255 categories achieved the best result in the pre-training phase. Although more images tend to be better in Table~\ref{tab:numof_image}, a huge 1M dataset does not necessarily lead to better results  in Section~\ref{sec:expolarative}.} 

\noindent \textbf{Other investigations (see also Figure~\ref{figD}, Tables~\ref{tab:line}, and Tables~\ref{tab:width}).} For investigations (F2) and (F5), these parameters achieved slightly higher mIoU, but had no significant effect on ADE-20k fine-tuning. A recent study~\cite{qin2022} claimed that a highly accurate annotation like  segmentation in ${m}^{1}$ is important, but it had slightly better accuracy than other configurations with a higher number of instances per image in SegRCDB pre-training.

\noindent \textbf{Comparison experiments.} Undoubtedly, the proposed SegRCDB pre-training model is more accurate than self-supervised learning with synthetic datasets such as GTA5, RCDB, and ExFractalDB for fine-tuning in indoor scenes (ADE-20k) and urban scenes (Cityscapes). Moreover, the SegRCDB pre-trained Swin Transformer performed equally well or even better than a sophisticated supervised learning method with semantic segmentation datasets. Actually, the proposed method exhibited better performance compared with that of COCO-Stuff pre-training. Among the FDSL methods, SegRCDB largely achieved better accuracy than those of ExFractalDB and RCDB. The FDSL method, specialized for classification problems, has been successfully applied to semantic segmentation tasks.

\noindent \textbf{Limitations.} We believe that there are additional factors other than (F1)--(F7) that have not been considered for improving segmentation pre-training.  It is also possible to examine the relationship between factors in detail. Finally, the shapes in our SegRCDB were taken "as is" from a previous study~\cite{Kataoka_2022_CVPR}. Therefore, there is a room to further improve our SegRCDB for semantic segmentation pre-training.

\begin{table}
  \centering
  \caption{Annotation analysis on shift and inflation. The values are given in pixels.}
  \begin{tabular}{lccccc}
    \toprule
     Pixels  & 0 & 10 & 30 & 100 & 300\\
    \midrule
       Shift & 70.23& 70.02 & 70.08 & 63.69 &16.04\\
       Inflation & 70.23 & 70.86 & 69.18 & 67.30 &  15.73 \\
    \midrule
    \begin{minipage}{.15\linewidth}
    \centering
    \includegraphics[scale=0.34]{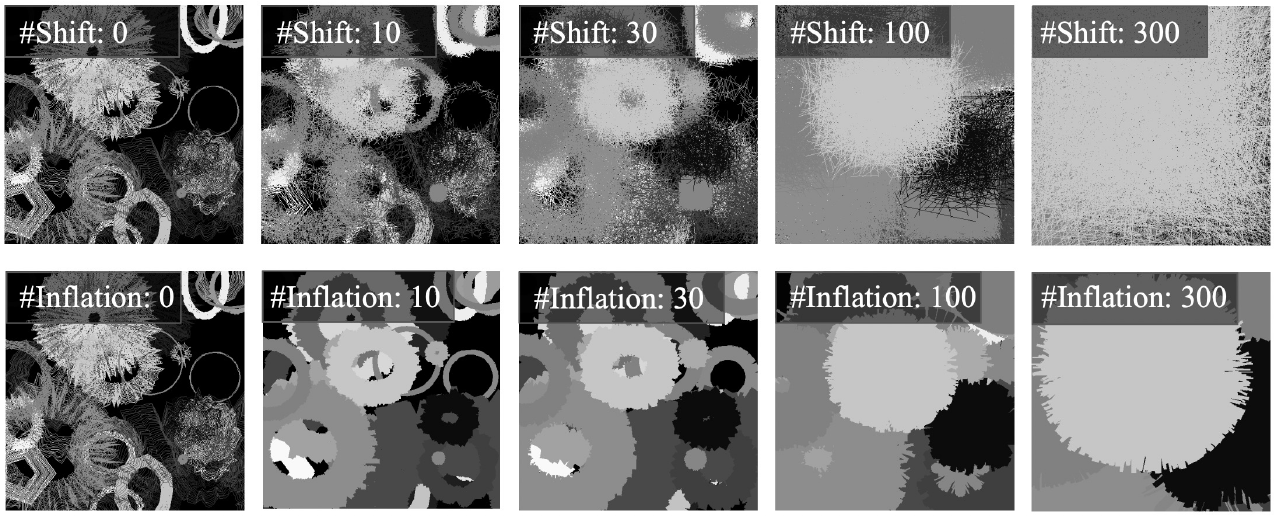}
    \end{minipage} \\ 
  \bottomrule
  \end{tabular}
  \vspace{-10pt}
  \label{tab:annotation_analysis}
\end{table}

\section{Conclusion}
We proposed SegRCDB (Segmentation Radial Contour DataBase), the first segmentation dataset developed by formula-driven supervised learning (FDSL). By investigating effective parameters for pre-training, SegRCDB outperformed COCO-Stuff-164k pre-training model with the same number of training data. 

SegRCDB can reveal what is effective in pre-training for semantic segmentation by varying the dataset's parameters. We hope that our SegRCDB will promote research on semantic segmentation pre-training without any manual effort and real images.

\section{Acknowledgement}
This paper is based on results obtained from a project,
JPNP20006, commissioned by the New Energy and Industrial Technology Development Organization (NEDO).
Computational resource of AI Bridging Cloud Infrastructure (ABCI) provided by National Institute of Advanced Industrial Science and Technology (AIST) was used. 

{\small
\bibliographystyle{ieee_fullname}
\bibliography{main}
}

\def\figtitle{
\begin{figure*}
\centering
\includegraphics[width=17cm]{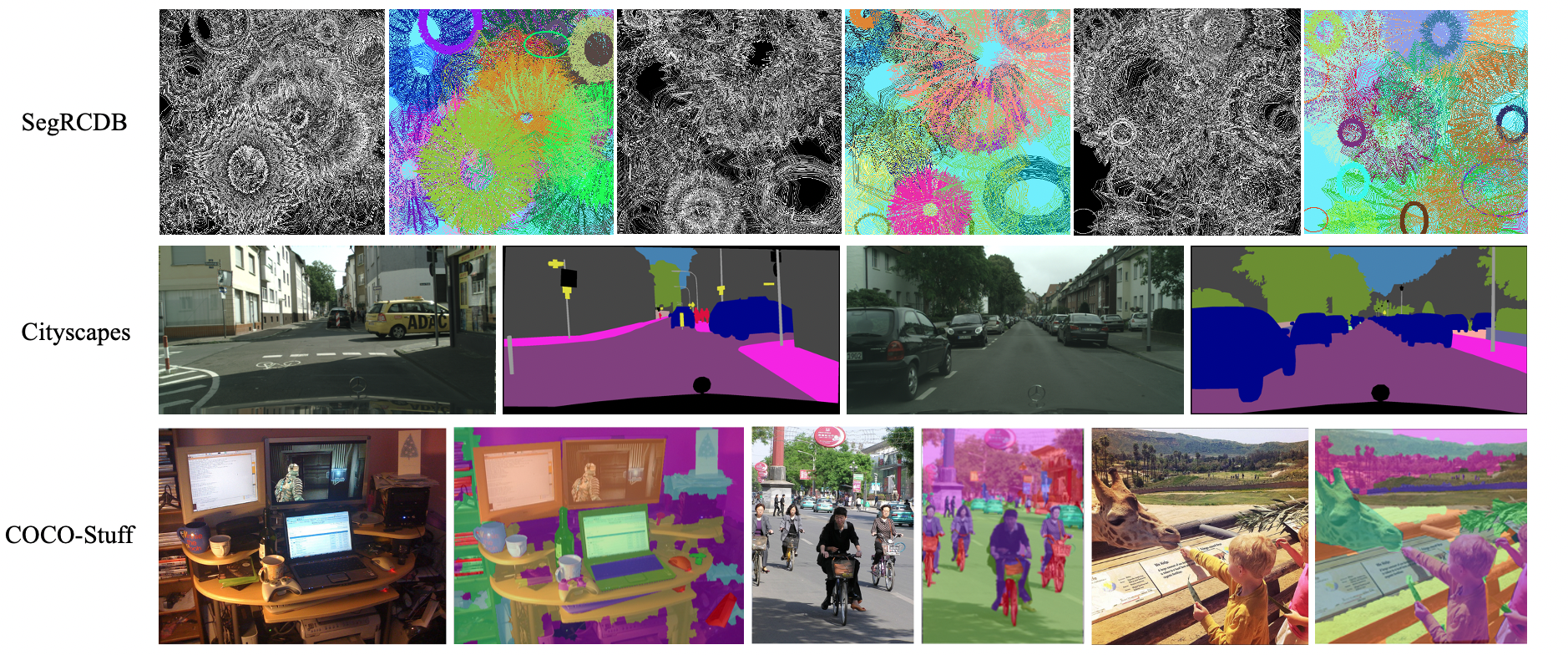}
\caption{\textbf{Example images of semantic segmentation datasets.}
    The first row is the visualization of SegRCDB.
    The second row and third row are the visualization of Cityscapes~\cite{Cordts2016Cityscapes} and COCO-Stuff~\cite{https://doi.org/10.48550/arxiv.1612.03716}.
    }
\label{fig:title}
\end{figure*}
}
\clearpage
\appendix 
\figtitle
\section{Dataset details}
\label{dataset}
We list the example images of SegRCDB and other datasets in Fig~\ref{fig:title}. SegRCDB generation code is available at: \url{https://github.com/dahlian00/SegRCDB}.

\section{Experiment Details}
\label{experiment_details}
Throughout this experiment, in the augmentation process, we adopt MMSegmentation~\cite{mmseg2020} official settings. We use the following augmentations: resize, random crop, random flip, photometric distortion, normalize, and padding.  

As the model architecture, we utilize UPerNet~\cite{xiao2018unified} as our base model with a Swin Transformer base~\cite{liu2021Swin} backbone. Throughout the paper, we adopt cross entropy loss with a loss weight of 1.0 for UPerNet head training.

\subsection{Investigation (Main paper Sec~5.2)}
\textbf{Settings}.
We set 300 epochs for pre-training and 60 epochs for fine-tuning.  In investigation (F1) - (F6), all parameters follow MMSegmentation official settings, with batch size set to 32. 
In (F7), fine-tuning is set to 120 epochs to allow for convergence. 
SegRCDB parameters in (F7) are set to the best result obtained from the investigation.

Table~\ref{SegRCDB_param} compares the base parameters in the investigation with the parameter combinations that gave the best results. 
Table~\ref{investigation_result} shows the result using the base parameters and best parameters of SegRCDB with 20k images fine-tuned on ADE-20k~\cite{zhou2017scene}. Fine-tuning epoch is set to 60 epochs. 
This confirms that SegRCDB has acquired more effective image representations due to the investigation, with a 20.51 mIoU improvement over the baseline.
\begin{table}
  \centering
  \caption{SegRCDB parameter}
  \begin{tabular}{lcc}
    \toprule
     &Base line &Best\\
    \midrule
       Number of instances ($N$) & $1$& $32$ \\
       Mask type & $\bm{m}^{1}$ & $\bm{m}^{1}$ \\
       Line width ($d$)& $1.0$  & $1.0$ \\
       Number of polygons ($K$)& \{$1$, $2$...$50$\}  & \{$1$, $2$...$25$\}\\
        Occlusion ($r$) & $512$ & $400$\\
      Colors & grayscale& grayscale\\
      Number of categories ($C$) & $255$ & $255$ \\
    \bottomrule
  \end{tabular}
  \label{SegRCDB_param}
  \vspace{-10pt}
\end{table}
\setlength\floatsep{20pt}
\begin{table}[t]
\begin{center}
    \caption{Investigation result at ADE-20k-val. SegRCDB contains 20k training images. Fine-tuning is set to 60 epochs. } 
    \begin{tabular}{lcc}
      \toprule
      & Base line& Best \\
       \midrule
        mIoU& 19.21 & 39.72\\
    \bottomrule
 \end{tabular}
\label{investigation_result}
\end{center}
\vspace{-10pt}
\end{table}

\subsection{Supervised pre-training for semantic segmentation
datasets. (Main paper Sec~5.3)}
\label{sec:semseg}

\textbf{Settings}.
For the pre-training of natural image datasets, we follow the MMSegmentation official settings, with the backbone learning rate set to 0.00024 and a batch size of 64.
SegRCDB can converge even when using a higher learning rate; therefore, we set a higher backbone learning rate of 0.0006. In the fine-tuning phase, the learning rate is set to 0.0005, and the batch size to 16 for all datasets. 

\subsection{Backbone pre-training for semantic segmentation. (Main paper Sec~5.3)}

\textbf{Settings}.
For ImageNet-1k, RCDB-1k~\cite{Kataoka_2022_CVPR}, and ExFractalDB-1k~\cite{Kataoka_2022_CVPR} training, we followed the official settings of Swin Transformer~\cite{liu2021Swin}, with the exception of the augmentation process, where we adopt MMSegmentation's official settings.  We use the following augmentations: resize, random crop, random flip, photometric distortion, normalize, and padding. SegRCDB pre-training settings are the same as Sec~\ref{sec:semseg}. Fine-tuning is 150 epochs long for all datasets. 

In the fine-tuning phase using ImageNet backbone, the learning rate is set to 0.00006 following MMSegmentation official settings. The backbone learning rate of RCDB and ExFractalDB is set to 0.001 to allow for convergence. 
We also tested a learning rate of 0.001 when using a pre-trained ImageNet backbone for fairness. Table~\ref{learningrate_backbone} shows the results of fine-tuning ImageNet with a backbone learning rate of 0.001.  The fine-tuning from ImageNet shows lower mIoU compared to the official setting (ADE-20k: 43.60 -- 46.37, Cityscapes: 71.85 --75.26). For this reason, in the main paper, we show follow official learning rates for ImageNet's fine-tuning phase, for the fairness comparison.

\begin{table*}[t]
\begin{center}
 \caption{Comparison with backbone pre-training. This represents the result of fine-tuning ImageNet with a learning rate of 0.001 as other classification datasets.} 
    \begin{tabular}{lccccc}
      \toprule
        &  &\multicolumn{2}{c}{ADE-20k} & \multicolumn{2}{c}{Cityscapes} \\
        Pre-training & \#Img &  mIoU & mAcc & mIoU & mAcc \\
        \bottomrule[0.5pt]
        Scratch & -& 31.40 & 41.02 & 54.65 & 62.89 \\
        ImageNet& 1.28M & 43.60& 54.66 &   71.85& 80.71 \\
        ExFractalDB&1M&  41.10 & 52.05  & 68.93 & 77.96\\
        RCDB&1M &  38.50 & 49.43 &  66.57 & 75.88  \\
        SegRCDB&118k & \textbf{43.85} & \textbf{54.98} & \textbf{73.06} & \textbf{81.59}    \\
        \bottomrule[0.8pt]
 \end{tabular}
\label{learningrate_backbone}
\end{center}
\vspace{-10pt}
\end{table*}

\section{Visualization}
\begin{figure*}
\centering
\includegraphics[width=18cm]{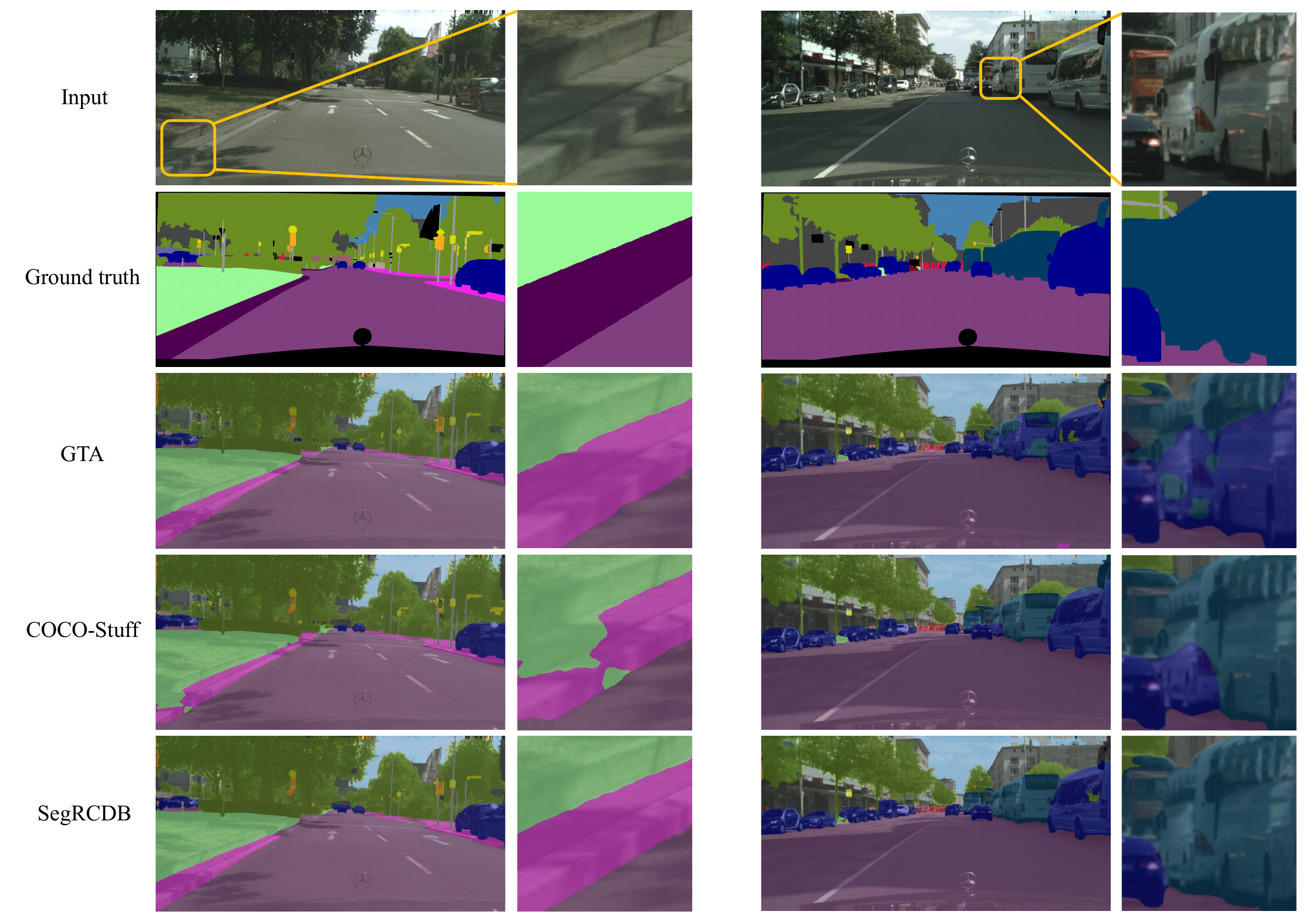}
\caption{Visual comparison of fine-tuning result on Cityscapes. The first low represents the input images, and the second row represents the ground truth. An enlarged detail of the image is attached on the right. SegRCDB used 118k images. }
\label{fig:city}
\vspace{-15pt}
\end{figure*}

\begin{figure*}
\centering
\includegraphics[width=18cm]{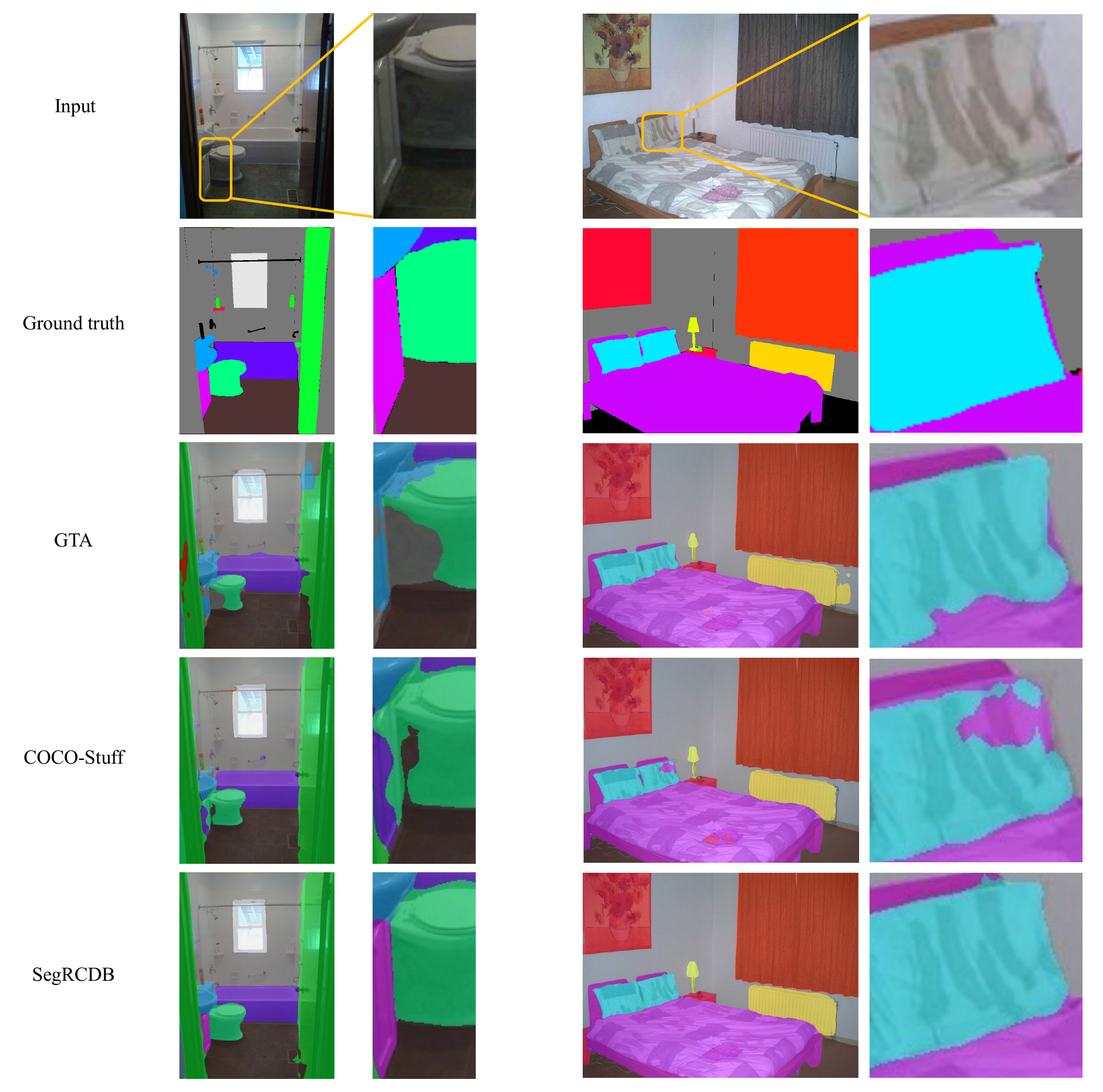}
\caption{Visual comparison of fine-tuning result on ADE-20k. The first low represents the input images, and the second row represents the ground truth. An enlarged detail of the image is attached on the right. SegRCDB used 118k images. }
\label{fig:ade}
\vspace{-15pt}
\end{figure*}

\label{sec:visualization}
Figure~\ref{fig:city} and Figure~\ref{fig:ade} show the visualizations of fine-tunings on Cityscapes and ADE-20k. As can be seen, models pre-trained on SegRCDB can still annotate details comparable to pre-trained models on other real image datasets, despite not having seen any natural images during training. 

\end{document}